\PassOptionsToPackage{dvipdfmx}{graphicx}
\documentclass[runningheads]{llncs}
\usepackage{graphicx}
\usepackage{amsmath}
\usepackage{mathtools}
\usepackage{bm}
\usepackage{amssymb}
\usepackage{graphicx}
\usepackage{xcolor}  
\usepackage{booktabs}
\usepackage{bbm}
\usepackage{multirow}
\usepackage[numbers, square, comma]{natbib}

\usepackage {colortbl,array,xcolor}
\usepackage{enumitem}



\begin{document}
\title{FLARE-SSM: Deep State Space Models \\ with Influence-Balanced Loss \\ for 72-Hour Solar Flare Prediction}
\titlerunning{72-Hour Solar Flare Prediction}
%
\author{Yusuke Takagi \and
Shunya Nagashima \and
Komei Sugiura}
\authorrunning{Y. Takagi et al.}
%
\institute{Keio University, Japan\\
\email{\{yusuke.10.06, ng\_sh, komei.sugiura\}@keio.jp}}
\maketitle              
\begin{abstract}

Accurate and reliable solar flare predictions are essential to mitigate potential impacts on critical infrastructure. However, the current performance of solar flare forecasting is insufficient.
In this study, we address the task of predicting the class of the largest solar flare expected to occur within the next 72 hours.
Existing methods often fail to adequately address the severe class imbalance across flare classes.
To address this issue, we propose a solar flare prediction model based on multiple deep state space models.
In addition, we introduce the frequency \& local-boundary-aware reliability loss (FLARE loss) to improve predictive performance and reliability under class imbalance.
Experiments were conducted on a multi-wavelength solar image dataset covering a full 11-year solar activity cycle.
As a result, our method outperformed baseline approaches in terms of both the Gandin–Murphy–Gerrity score and the true skill statistic, which are standard metrics in terms of the performance and reliability.

\keywords{AI for Science \and Solar Flare Prediction \and Class Imbalance \and Deep SSM \and Imbalanced Data Learning.}
\end{abstract}

\section{Introduction}

Solar flares are phenomena that can severely impact critical infrastructure, including GPS systems, communication networks, spacecraft, and power grids~\cite{Bhattacharjee_ApJ_2020, Deshmukh_aaai_2021}.
In particular, Carrington-class flares~\cite{Cliver_2013} have the potential to cause economic damage estimated as ranging from 0.6 to 2.6 trillion USD~\cite{Maynard_techreport_2013}.
Given this context, accurate and reliable solar flare predictions are of paramount importance.
Nevertheless, current solar flare forecasting remains a highly challenging task with limited levels of performance.


In this study, we address the task of predicting the class of the largest solar flare expected to occur within the next 72 hours.
Typical use cases of such predictions include implementing countermeasures such as rerouting or rescheduling of flights, protection of power grids, safe-mode transitions of satellites, and error correction in positioning systems.


Predicting the maximum class of solar flare within a 72-hour window is a highly challenging task.
Indeed, in terms of the Gandin-Murphy-Gerrity score (GMGS)~\cite{Gandin_gmgs_1992} and the Brier skill score (BSS)~\cite{Nishizuka_bss_2020}, expert-based 24-hour forecasts during the period from 2010 to 2015 achieved only $\text{BSS}_{\ge \text{M}} = 0.16$ and $\text{GMGS} = 0.48$~\cite{Kubo_humanperf_2017, Murray_humanperf_2017}.
Considering that expert performance is limited even for 24-hour forecasts, predicting the maximum flare class within a 72-hour horizon is expected to be even more difficult.
While recent methods for solar flare prediction~\cite{Nagashima_2025, Sun_ApJ_2022} have reported promising results for 24-hour forecasts, they often lack adequate mechanisms to address the severe class imbalance existing across flare classes.

\begin{figure}[t]
    \vspace{-2mm}
    \centering
    \includegraphics[width=\linewidth]{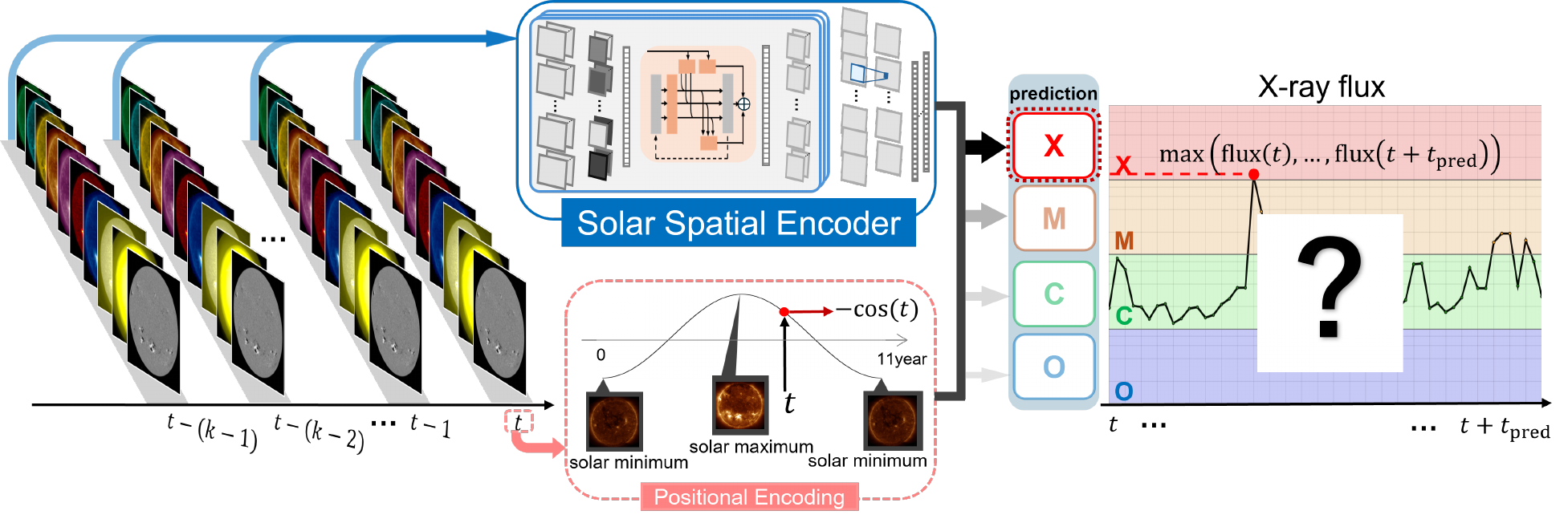}
    \vspace{-7mm}
    \caption{Overview of the proposed method. The input consists of a sequence of multi-wavelength solar images captured by Atmospheric Imaging Assembly (AIA) and the Helioseismic and Magnetic Imager (HMI) instruments (top), which are processed by the Solar Spatial Encoder to extract spatio-temporal features. The timestamp (bottom) is encoded into a scalar value based on the 11-year solar activity cycle. These two representations are then integrated to predict the probability distribution over four solar flare classes (X, M, C, and O).}
    \label{fig:eyecatch}
\end{figure}

To address this challenge, we propose a solar flare prediction model based on deep state space models.
We introduce the frequency \& local-boundary-aware reliability loss (FLARE loss) to improve both the predictive performance and the forecast reliability under severe class imbalance.
The main differences in our method with respect to existing methods are the use of the FLARE loss, which applies sample-wise weighting based on class frequency and influence, and the incorporation of input time embeddings based on the 11-year solar activity cycle.
The FLARE loss is expected to enhance classification performance under class imbalance and mitigate overfitting to samples near decision boundaries as training progresses.
Moreover, explicitly providing the model with the phase information of the input time within the solar cycle through positional embeddings results in a further improvement in the predictive performance. An overview of the proposed method is illustrated in Figure~\ref{fig:eyecatch}.

The main contributions of this work are as follows.
\begin{itemize}
    \item We introduce the FLARE loss, which consists of the following three components: 
    (i) the influence-balanced (IB) loss~\cite{Park_ICCV_2021}, which suppresses the excessive influence of samples near decision boundaries; (ii) the IB BSS loss, which incorporates influence-based weighting~\cite{Park_ICCV_2021} into the BSS loss~\cite{Kaneda_FlareTransformer_2022}; and (iii) the Weighted BSS loss, which introduces class-wise weighting into the BSS loss.     
    \item We perform 72-hour solar flare prediction by incorporating input time embeddings derived from the 11-year solar activity cycle.
\end{itemize}


\section{Related Work}

Numerous studies have been conducted on solar flare prediction~\cite{Jonas_2018, Kusano_Science_2020, Nagashima_2025, Nishizuka_ApJ_2017}, and Georgoulis et al.~\cite{Georgoulis_JSWSC_2021} provide a comprehensive overview of this research field.
However, a significant challenge in this area is the severe imbalance in the frequency of occurrence across solar flare classes, making it essential to address this issue.
Various approaches have been proposed to tackle class imbalance~\cite{Berthelot_2019, Chawla_2002, Cui_2019, Lin_2017, Park_ICCV_2021}, and an overview is provided by Krawczyk et al.~\cite{Krawczyk_2016}.


A wide range of methods have been proposed for solar flare prediction~\cite{Deshmukh_aaai_2021, Iida_ACCV_2022, Kaneda_FlareTransformer_2022, Li_AAS_2024, Liu_ApJ_2019, Nagashima_2025, Park_ApJ_2018}.
Among them, several approaches leverage Convolutional Neural Networks (CNNs), Recurrent Neural Networks, and Long Short-Term Memory (LSTM) networks, which are well-suited for processing images and image sequences~\cite{Bhattacharjee_ApJ_2020, Li_AAS_2020, Liu_ApJ_2019, Park_ApJ_2018}.
Liu et al.~\cite{Liu_ApJ_2019} proposed an LSTM model that incorporates both magnetic field parameters and the flare history to capture the temporal evolution of active regions, demonstrating the importance of modeling temporal dynamics.
More recently, Transformer-based models have been explored because of their superior ability to capture long-range dependencies~\cite{Abduallah_SolarFlareNet_2023, Grim_SoPh_2024, Iida_ACCV_2022, Kaneda_FlareTransformer_2022, Li_AAS_2024}.
Flare Transformer~\cite{Kaneda_FlareTransformer_2022} introduces a method that applies the attention mechanism of Transformers to sunspot and magnetic field features, modeling the temporal relationships between images and physical parameters.
Furthermore, Nagashima et al.~\cite{Nagashima_2025} proposed the Deep Space Weather Model (Deep SWM), which uses only multi-channel solar images as input and performs 24-hour solar flare prediction using a masked autoencoder and a deep state space model.


Research on class imbalance can be broadly categorized into three main approaches: (i) data-level approaches; (ii) cost-sensitive re-weighting approaches; and (iii) meta-learning approaches.
(i) Data-level approaches aim to adjust the class distribution in the training set. Representative methods include SMOTE~\cite{Chawla_2002}, which synthesizes samples for minority classes, and ReMixMatch~\cite{Berthelot_2019}, which dynamically balances class distributions within training batches.
(ii) Cost-sensitive re-weighting approaches assign different weights during loss computation. Examples include Focal Loss~\cite{Lin_2017}, which emphasizes hard-to-classify samples; Class-Balanced Loss~\cite{Cui_2019}, which weights samples based on the effective number of instances; and IB Loss~\cite{Park_ICCV_2021}, which incorporates sample-wise influence into the loss via inverse weighting.
(iii) Meta-learning approaches aim to learn the weighting function or loss design itself. One notable example is Meta-Weight-Net~\cite{Shu_2019}, which learns the weighting function from external data.


In solar flare prediction, several standard datasets have been established for different tasks.
Nishizuka et al.~\cite{Nishizuka_ApJ_2017} proposed a dataset based on physical features of sunspots extracted from images captured by the Solar Dynamics Observatory (SDO)~\cite{Pesnell_SDO_2012} and the Geostationary Operational Environmental Satellite (GOES), covering the period from June 2010 to December 2015.
Angryk et al.~\cite{Angryk_ScientificData_2020} constructed a dataset that includes physical features of active regions extracted from the Spaceweather HMI Active Region Patch (SHARP) series, spanning from May 2010 to December 2018.
Furthermore, FlareBench~\cite{Nagashima_2025} was developed as a benchmark for full-disk image-based prediction tasks. It is constructed using observations from SDO's Helioseismic and Magnetic Imager (HMI) and Atmospheric Imaging Assembly (AIA) instruments and contains multi-wavelength image sequences spanning 11.5 years from June 2011 to November 2022.


Deep SWM is a model that takes as input 10-channel solar images spanning multiple time steps and predicts the class of the largest solar flare expected to occur within the next 24 hours.
It employs a two-stage training strategy, combining pretraining with a masked autoencoder~\cite{He_MAE_2022} and classifier re-training (cRT)~\cite{Kang_CRT_2020}.
In contrast, the method proposed in this study performs single-stage training using the FLARE loss, without relying on pretraining or cRT.
Moreover, our method differs in that it incorporates not only solar images but also temporal information as input, and it targets 72-hour flare prediction rather than 24-hour prediction.

\section{Problem Statement}

In this study, we address the task of predicting the class of the largest solar flare expected to occur within the next 72 hours.
We formulate this as a multi-class classification problem.
Specifically, solar flares are categorized into four classes (X, M, C, and O) based on the peak X-ray flux (i.e., the number of photons per unit time) observed during the prediction window.
This formulation is consistent with established practices in the field of solar flare forecasting~\cite{Leka_ApJ_2019, Nishizuka_ApJ_2017, Zhang_ASR_2024, Zheng_ApJ_2019}.

In this task, it is desirable to achieve accurate solar flare classification.
\begin{figure}[t]
    \vspace{-2mm}
    \centering
    \includegraphics[width=\linewidth]{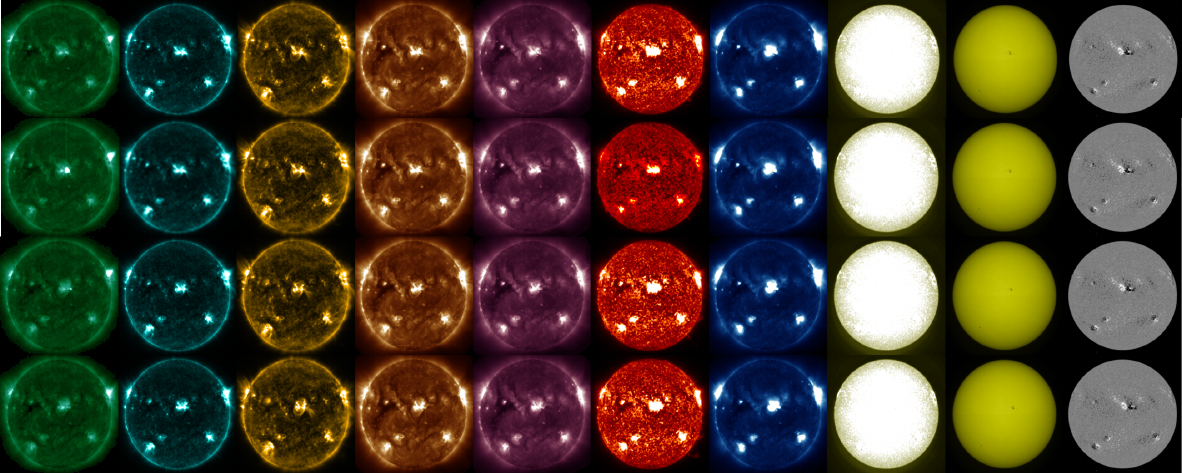}
    \vspace{-7mm}
    \caption{HMI and AIA images sampled every two hours from 00:00 to 06:00 on September 6, 2011. Horizontal and vertical axes correspond to channels and time, respectively.}
    \label{fig:typical_scene}
\end{figure}
Figure~\ref{fig:typical_scene} shows a representative example of the input used in this study: HMI and AIA images captured every two hours from 00:00 to 06:00 on September 6, 2011.
In this case, an X-class flare occurred within the following 72 hours and, therefore, the correct prediction should be the X class.


The input consists of ten types of images provided as time series: one type from the HMI ~\cite{Scherrer_SoPh_2012} instrument and nine types from the AIA ~\cite{Lemen_SoPh_2012} instrument, each corresponding to a different wavelength.
The output is the predicted probability for each of the four solar flare classes.
HMI images refer to solar images taken by the HMI instrument onboard NASA's SDO ~\cite{Pesnell_SDO_2012}, which captures the line-of-sight component of the magnetic field, providing essential magnetic information for solar flare prediction.
AIA images are obtained from a separate instrument also onboard the SDO and provide information concerning the solar atmosphere at multiple wavelengths.


Given the severe class imbalance across flare categories, trivial predictions—such as always classifying all events as O-class—are not meaningful.
Therefore, the goal is to make predictions that maximize standard evaluation metrics widely used in solar flare prediction, namely the GMGS ~\cite{Gandin_gmgs_1992} and the BSS~\cite{Nishizuka_bss_2020}.
For evaluation, we employ $\text{BSS}_{\ge \text{M}}$, $\text{TSS}_{\ge \text{M}}$~\cite{Kubo_humanperf_2017}, and GMGS as performance metrics, where TSS is the true skill statistic.
For BSS and TSS, the ``$\ge$M'' notation indicates that these metrics are applied after categorizing the output into two classes (``$\ge$M'' and ``$<$M'').

\section{Proposed Method}

In this study, we extend the Deep Space Weather Model~\cite{Nagashima_2025} to predict the class of the largest solar flare expected to occur within the next 72 hours.
To improve the reliability of the predicted probabilities under severe class imbalance, we introduce two loss functions that apply sample-wise weighting: the Weighted BSS loss and the IB BSS loss.
The use of Weighted BSS loss and IB BSS loss is considered broadly applicable to tasks with class imbalance, especially in cases where maintaining the reliability of the predicted probabilities is critical.


The main novelties of the proposed method are as follows:
\begin{itemize}
    \item We introduce the FLARE loss, which consists of the following:  
    (i) the IB BSS loss, which incorporates influence-based weighting~\cite{Park_ICCV_2021} into the BSS loss~\cite{Kaneda_FlareTransformer_2022};  
    (ii) the Weighted BSS loss, which introduces class-wise weighting into the BSS loss; and  
    (iii) the IB loss, which suppresses the excessive influence of samples near decision boundaries.
    
    \item We perform 72-hour solar flare prediction using input time embeddings based on the 11-year solar activity cycle.
\end{itemize}

\vspace{-5mm}
\subsection{Architecture}


\begin{figure*}[t]
    \vspace{-2mm}
    \centering
    \includegraphics[width=\linewidth]{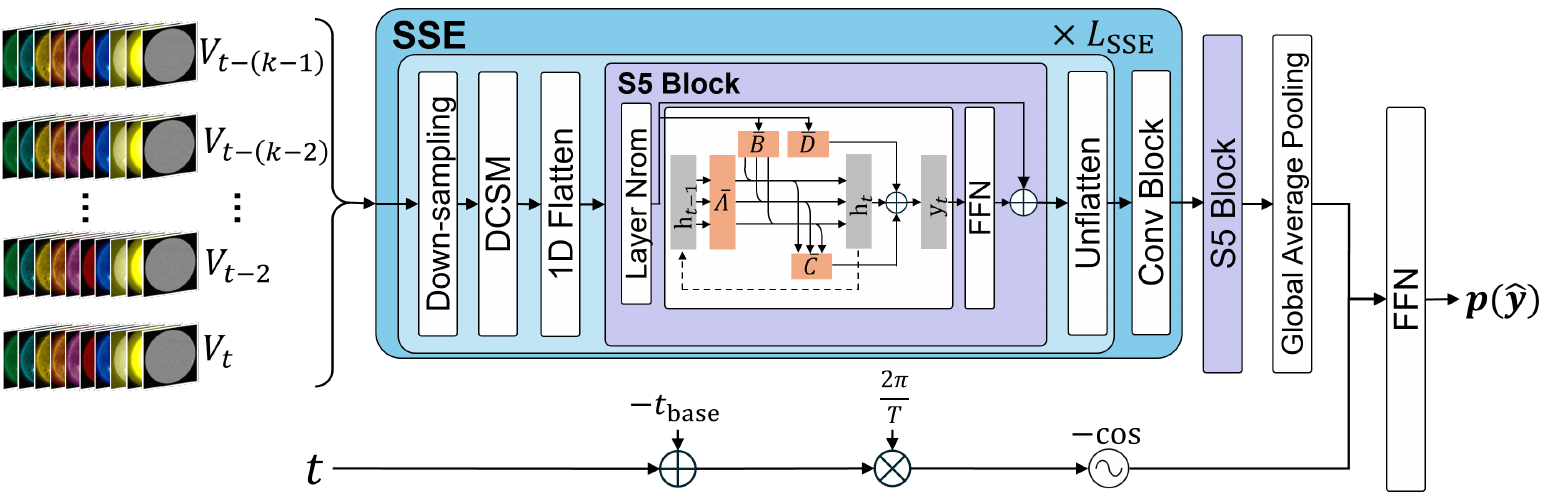}
    \vspace{-7mm}
    \caption{Architecture of the proposed method. The Solar Spatial Encoder (SSE) extracts spatio-temporal features from multi-wavelength AIA and HMI image sequences. In parallel, the timestamp is transformed into a positional embedding that reflects the phase of the 11-year solar activity cycle.}
    \label{fig:model}
\end{figure*}

Figure~\ref{fig:model} illustrates the model architecture of the proposed method.
The core module of our method is the Solar Spatial Encoder (SSE)~\cite{Nagashima_2025}.

We define the input image sequence as
$
\bm{x} = \left( \bm{V}_{t-(k-1)}, \bm{V}_{t-(k-2)}, \dots, \bm{V}_t \right) \in \mathbb{R}^{k \times C \times H \times W}$,
where $\bm{V}_t \in \mathbb{R}^{C \times H \times W}$ denotes a $C$-channel image at time $t$, and $k$, $H$, and $W$ represent the history length, image height, and image width, respectively.
Each channel in $\bm{V}_t$ corresponds to either an HMI image or an AIA image at a specific wavelength.
In addition to the image sequence, the most recent time step $t$ in the sequence is also provided as input to the model.

The SSE efficiently extracts long-range spatio-temporal dependencies from multi-channel solar image sequences while dynamically weighting the importance of each channel.
This module enables the model to capture fine-grained variations in active regions and characteristic patterns around sunspots, which is expected to improve the accuracy of solar flare prediction.

In addition, solar activity exhibits an approximately 11-year cycle, during which high-intensity flares occur more frequently~\cite{Hathaway_SolarActivityCycle_2010}.
By explicitly incorporating the phase information of the input time within this solar cycle, the model can better predict rare flare classes with low occurrence frequency.
The temporal embedding $\phi$ corresponding to the input time $t$ is defined as
\begin{equation}
    \phi = -\cos \left( 2\pi \frac{t - t_{\mathrm{base}}}{T} \right),
\end{equation}
where $t_{\mathrm{base}}$ and $T$ denote the start time and the period of the solar activity cycle, respectively.

Using $\bm{h}_{\text{SSE}}$ extracted by the SSE and $\phi$, the predicted probability $\bm{p}(\hat{\bm{y}})$ of the solar flare class given input $\bm{x}$ and time $t$ is computed as
\begin{equation}
    \bm{p}(\hat{\bm{y}}) = \text{FFN} \left[ \text{SSMBlock} \left( \bm{h}_{\text{SSE}} \right) ; \phi \right],
\end{equation}
where $\text{SSMBlock}$ follows the formulation proposed by Nagashima et al.~\cite{Nagashima_2025} and $\text{FFN}$ denotes a feedforward network.

\vspace{-2.5mm}
\subsection{Loss Function}

\subsubsection{IB loss}
The IB loss~\cite{Park_ICCV_2021} is a loss function designed to mitigate overfitting to the decision boundary under class imbalance by applying sample-wise weighting based on influence functions.

In the proposed method, let $\bm{h} \in \mathbb{R}^L$ denote the output of the final hidden layer and let $\bm{W}$ be the weight matrix of the fully connected layer that follows $\bm{h}$.
The predicted probability is then given by
\begin{equation}
    \bm{p}(\hat{\bm{y}}) = \text{softmax}(\bm{W} \bm{h}),
    \label{eq:pred}
\end{equation}
where $L$ is the number of nodes in the final hidden layer.

The IB loss based on the cross-entropy loss is defined as
\begin{equation}
    L^{\mathrm{IB}}_{\mathrm{CE}} = \frac{1}{|\mathcal{B}|} \sum_{(x, y) \in \mathcal{B}} \gamma(\bm{y}) \frac{L_{\mathrm{CE}}(\bm{y} , \bm{p}(\hat{\bm{y}}))}{\left\lVert \bm{p}(\hat{\bm{y}}) \right\rVert_1 \left\lVert \bm{h} \right\rVert_1} ,
\end{equation}
where $L_{\mathrm{CE}}$ denotes the cross-entropy loss for a single sample, $\bm{y}$ is the one-hot vector of the ground-truth label, and $\mathcal{B}$ is the mini-batch.
The weighting term $\gamma(\bm{y})$ is inversely proportional to the number of samples belonging to the class of the ground-truth label.

\subsubsection{Weighted BSS loss}
The Weighted BSS loss is a variant of the BSS loss~\cite{Nishizuka_bss_2020} in which class-wise weights inversely proportional to the number of samples are applied.
While the original BSS loss is effective in tasks where the reliability of predicted probabilities is important, it suffers from a dominance of loss contributions from majority classes, which can degrade the performance for minority classes.

To address this issue, we introduce the Weighted BSS loss in this study.
The BSS loss for a single sample is defined as
\begin{equation}
    L_{\mathrm{BSS}}(\bm{y}, \bm{p}(\hat{\bm{y}})) = \sum_{k=1}^K \left( p(\hat{y}_k) - y_k \right)^2 .
    \label{eq:BSS}
\end{equation}
Based on Equation~\eqref{eq:BSS}, the Weighted BSS loss is defined as
\begin{equation}
    L^{\mathrm{'}}_{\mathrm{BSS}} = \frac{1}{|\mathcal{B}|} \sum_{(x, y) \in \mathcal{B}} \gamma(\bm{y}) L_{\mathrm{BSS}}(\bm{y} , \bm{p}(\hat{\bm{y}})),
\end{equation}
where $\gamma(\bm{y})$ is the weight assigned to each sample based on the inverse frequency of its ground-truth class.

\subsubsection{IB BSS loss}
The IB BSS loss is a loss function that applies sample-wise weighting based on influence functions, in a manner similar to Park et al.~\cite{Park_ICCV_2021} but applied to the BSS loss.
While the BSS loss is effective for tasks that prioritize the reliability of the predicted probabilities, it suffers from a tendency to overfit as training progresses, because predictions are driven closer to 1 or 0 quadratically, making the model susceptible to overconfidence errors.

To mitigate this issue and maintain reliable predictions, we introduce the IB BSS loss, which suppresses the contribution of high-influence samples in the BSS loss.
Assuming the prediction follows Equation~\eqref{eq:pred} and that BSS loss is used instead of the cross-entropy, we derive the influence-aware weighting as follows.

First, the partial derivative of the BSS loss with respect to the $(k,l)$-th element of the final fully connected layer weight $\bm{W}$ is
\begin{equation}
    \frac{\partial L_{\mathrm{BSS}}}{\partial w_{kl}} = 2 h_l p(\hat{y}_k) \Big\{ \mathrm{\Delta}_k - \sum_{j=1}^N \mathrm{\Delta}_j p(\hat{y}_j) \Big\},
\end{equation}
where $\mathrm{\Delta}_k = p(\hat{y}_k) - y_k$.

Following the procedure of Park et al.~\cite{Park_ICCV_2021}, the IB weighting factor can be computed as
\begin{equation}
\begin{aligned}
    \sum_{k=1}^K \sum_{l=1}^L \left| \frac{\partial L_{\mathrm{BSS}}}{\partial w_{kl}} \right| 
    &= 2 \sum_{k=1}^K \Big| p(\hat{y}_k) \Big( \mathrm{\Delta}_k - \sum_{j=1}^N \mathrm{\Delta}_j p(\hat{y}_j) \Big) \Big| \sum_{l=1}^L \left| h_l \right| \\
    &= 2 \sum_{k=1}^K \left| p(\hat{y}_k) \Big( \mathrm{\Delta}_k - \bm{\mathrm{\Delta}} \cdot \bm{p}(\hat{\bm{y}}) \Big) \right| \sum_{l=1}^L \left| h_l \right| \\
    &= 2 \left\lVert \bm{p}(\hat{\bm{y}}) \odot \left\{ \bm{\mathrm{\Delta}} - \bm{1}  \left( \bm{\mathrm{\Delta}} \cdot \bm{p}(\hat{\bm{y}}) \right) \right\} \right\rVert_1 \left\rVert \bm{h} \right\rVert_1 ,
\end{aligned}
\end{equation}
where $\bm{\mathrm{\Delta}} = \bm{p}(\hat{\bm{y}}) - \bm{y}$.

Using this result, the IB BSS loss is defined as
\begin{equation}
    L^{\mathrm{IB}}_{\mathrm{BSS}} = 
    \frac{1}{|\mathcal{B}|} \sum_{(x, y) \in \mathcal{B}} \gamma(\bm{y}) 
    \frac{L_{\mathrm{BSS}}(\bm{y} , \bm{p}(\hat{\bm{y}}))}{2 \left\lVert \bm{p}(\hat{\bm{y}}) \odot \left\{ \bm{\mathrm{\Delta}} - \bm{1} \left( \bm{\mathrm{\Delta}} \cdot \bm{p}(\hat{\bm{y}}) \right) \right\} \right\rVert_1 \left\lVert \bm{h} \right\rVert_1} ,
\end{equation}
where $\odot$ denotes the Hadamard product and $\bm{1}$ is a vector of all ones with the same shape as $\bm{y}$.

\subsubsection{FLARE loss}
To improve the classification performance under class imbalance and to suppress overfitting to samples near the decision boundary as training progresses, we adopt the following composite loss function,
\begin{equation}
    L = \left( L^{\mathrm{'}}_{\mathrm{CE}} + L^{\mathrm{IB}}_{\mathrm{CE}} \right) + \lambda_{\mathrm{BSS}} \left( L^{\mathrm{'}}_{\mathrm{BSS}} + L^{\mathrm{IB}}_{\mathrm{BSS}} \right),
\end{equation}
where $L^{\mathrm{'}}_{\mathrm{CE}}$ denotes the Weighted Cross-Entropy (CE) loss and $\lambda_{\mathrm{BSS}}$ is a hyperparameter that balances the contribution of BSS-based losses relative to CE-based losses.
For training stability, the IB loss and IB BSS loss are not applied during the initial phase of training.

\section{Experiment Setup}

In this study, we used FlareBench~\cite{Nagashima_2025} for the model training and evaluation.
This benchmark covers a wide range of solar activity phases over an 11-year cycle, enabling training, validation, and testing without bias toward any particular period of solar activity.
While the original ground-truth labels in FlareBench correspond to the maximum flare class within the next 24 hours, we modified the task to predict the maximum flare class within the next 72 hours.
In addition, we constructed a dataset with 2-hour intervals by excluding samples corresponding to odd time steps.


The dataset consists of 56,616 samples collected over approximately 13 years, from June 2011 to April 2024.
Each sample contains 10 channels: one HMI image and nine AIA images corresponding to different wavelengths.
The samples are temporally spaced at 2-hour intervals.
We excluded 8,721 samples because of missing class labels or missing images in at least 25\% of the channels.
As a result, 47,895 samples were used for model training and evaluation.
For samples with less than 25\% missing image channels, the missing channels were replaced with images in which all pixel values were set to zero.
The numbers of samples corresponding to X-, M-, C-, and O-class flares were 2,131, 10,986, 16,608, and 18,170, respectively.
Following the protocol of Nagashima et al.~\cite{Nagashima_2025}, we split the dataset into training, validation, and test sets using time-series cross-validation.
The numbers of samples in the training, validation, and test sets were 31,085, 4,107, and 8,386, respectively.
The training set was used to update the model parameters, the validation set was used for hyperparameter tuning, and the test set was used for the final evaluation.


The experimental settings used in the proposed method are summarized in Table~\ref{tab:experiment_settings}.
\begin{table}[tb]
  \centering
  \caption{Experimental settings for the proposed method.}
  \label{tab:experiment_settings}
  \begin{tabular}{ll}
    \hline
    Epoch & 20 \\
    Batch size & 64 \\
    Optimizer & AdamW($\beta_1 = 0.9, \beta_2 = 0.95$) \\
    Learning rate & $4.0\times 10^{-5}$ \\
    Weight decay & $5.0\times 10^{-2}$ \\
    $\lambda_{\text{BSS}}$ & $3.0$ \\
    $t_{\text{base}}$ & 2008/12/01 00:00 UTC \\
    $T$ & 48,204 \\
    \hline
  \end{tabular}
\end{table}
The total number of trainable parameters was approximately 242 million, and the number of multiply–accumulate operations (MACs) was approximately 2.1 billion.
Training was conducted using a single GeForce RTX 4090 GPU with 24 GB of memory and an Intel Core i9-14900F CPU.
The total training time was approximately 1.5 hours, and the inference time per sample was approximately 11 ms.
During training, the GMGS score was computed on the validation set at each epoch.
For the final evaluation on the test set, we used the model checkpoint that achieved the highest GMGS score on the validation set.

\section{Results and Discussion}
\subsection{Quantitative Results}

\begin{table}[t]
  \centering
  \caption{Quantitative comparison. The best scores are in bold.
}
  \label{tab:quantitive}
  \setlength{\tabcolsep}{10pt}
  \begin{tabular}{llll}
    \hline
    Method & $\text{GMGS}\uparrow$ & $\text{BSS}_{\ge \text{M}}\uparrow$ & $\text{TSS}_{\ge \text{M}}\uparrow$ \\
    \hline \hline
    CNN-LSTM & $0.359 \pm 0.036$ & $\bm{0.434} \pm 0.123$ & $0.380 \pm 0.095$ \\
    Deep SWM\cite{Nagashima_2025} & $0.418 \pm 0.085$ & $0.024 \pm 0.275$ & $0.409 \pm 0.142$ \\
    Ours & $\bm{0.484} \pm 0.084$ & $0.353 \pm 0.089$ & $\bm{0.447} \pm 0.127$ \\
    \hline
  \end{tabular}
\end{table}

Table~\ref{tab:quantitive} presents a quantitative comparison between the proposed method and the baseline methods.
Each metric is reported as the mean and standard deviation computed over the first, second, and third folds in the time-series cross-validation.


We used CNN-LSTM and the Deep SWM ~\cite{Nagashima_2025} as baseline methods.
In solar flare prediction, approaches that combine CNNs with LSTM networks have been widely adopted~\cite{Nishizuka_DeFN_2018, Sun_ApJ_2022}.
Therefore, we selected a CNN-LSTM baseline similar to the model proposed by Sun et al.~\cite{Sun_ApJ_2022}.
Deep SWM was chosen because it has demonstrated strong performance on the task of predicting the maximum flare class within 24 hours using 10-channel solar images composed of HMI and AIA data.


We adopted $\text{GMGS}$~\cite{Gandin_gmgs_1992}, $\text{BSS}_{\ge \text{M}}$~\cite{Nishizuka_bss_2020}, and $\text{TSS}_{\ge \text{M}}$~\cite{Kubo_humanperf_2017} as evaluation metrics.
These are widely used standard metrics for this task~\cite{Leka_ApJ_2019, Nishizuka_ApJ_2017, Zhang_ASR_2024, Zheng_ApJ_2019}.
Specifically, GMGS ensures a fair evaluation across all flare classes.  
$\text{BSS}_{\ge \text{M}}$ assesses the forecast reliability for larger ($\ge$ M) and smaller ($<$M) flares.  
$\text{TSS}_{\ge \text{M}}$ balances the accurate prediction of both larger and smaller flares.


As shown in Table~\ref{tab:quantitive}, the proposed method achieved a GMGS score of 0.484, compared with 0.359 for CNN-LSTM and 0.418 for Deep SWM.
This corresponds to improvements of 0.125 points over CNN-LSTM and 0.066 points over Deep SWM.
For $\text{TSS}_{\ge \text{M}}$, the proposed method achieved a score of 0.447, while CNN-LSTM and Deep SWM scored 0.380 and 0.409, respectively.
This reflects improvements of 0.067 points over CNN-LSTM and 0.038 points over Deep SWM.
These results demonstrate that the proposed method outperforms both baseline methods.
The improvement in GMGS over the baselines was statistically significant ($p < 0.05$).

\subsection{Qualitative Results}

\begin{figure}[bt]
    \vspace{-3mm}
    \centering
    \includegraphics[width=\linewidth]{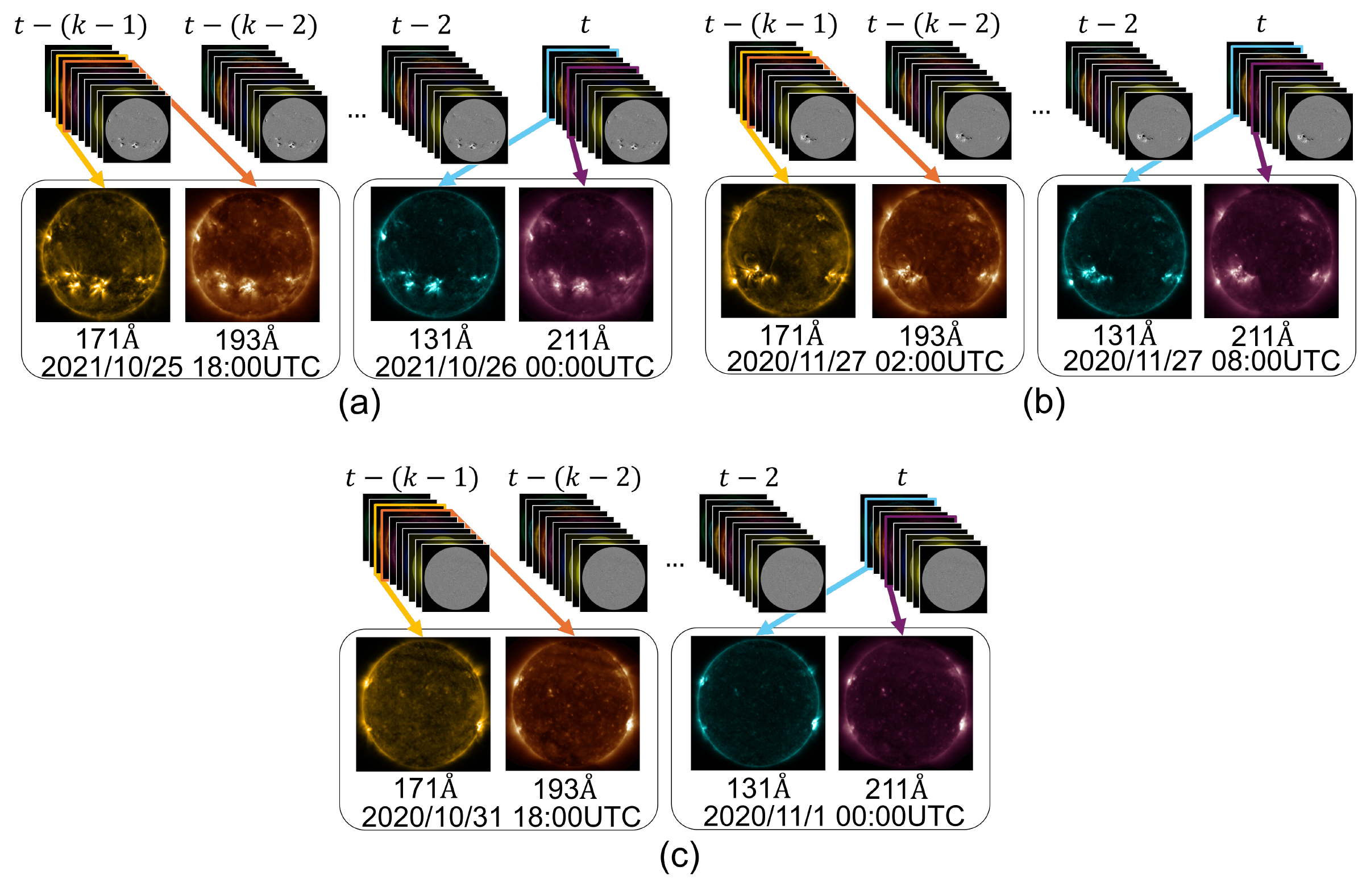}
    \vspace{-7mm}
    \caption{Qualitative results of the proposed method. Subfigures (a) and (b) show successful predictions, while subfigure (c) shows a failed prediction. Each subfigure shows AIA images at 171 Å and 193 Å taken at time $t-(k-1)$, and images at 131 Å and 211 Å taken at time $t$.}
    \label{fig:qualitative}
\end{figure}

Figure~\ref{fig:qualitative} presents qualitative examples from the proposed method.  
Subfigures (a) and (b) show cases in which the model correctly predicted X-class and M-class flares, respectively, while subfigure (c) shows a failure case.
Each subfigure displays AIA images at 171 Å and 193 Å taken at time $t-(k-1)$, and images at 131 Å and 211 Å taken at time $t$.
In subfigure (a), the input images are from 00:00 UTC on October 26, 2021. The model predicted an X-class flare within 72 hours, and an X-class flare was indeed observed approximately 63 hours later, indicating a correct prediction.
In subfigure (b), the input corresponds to 08:00 UTC on November 27, 2020. The model predicted an M-class flare, and an M-class flare was observed approximately 53 hours later, also indicating a correct prediction.
In contrast, subfigure (c) shows a case from 00:00 UTC on November 1, 2020. Although the maximum flare observed within 72 hours was a C-class flare (approximately 17 hours later), the model incorrectly predicted an X-class flare. This misprediction is likely due to the presence of multiple active regions near the solar limb, which the model may have mistakenly interpreted as precursors to a strong flare.

\subsection{Ablation Studies}

\begin{table}[bt]
  \centering
  \caption{Results of the ablation study. PE denotes the positional embedding, and HM represents the harmonic mean of $\text{GMGS}$ and $\text{BSS}_{\ge \text{M}}$. The best scores are in bold.}
  \label{tab:ablation}
      \setlength{\tabcolsep}{6pt}
      \begin{tabular}{ccccccccc}
        \hline
        Model & PE & $L^{\text{IB}}_{\text{CE}}$ & $L^{'}_{\text{BSS}}$ & $L^{\text{IB}}_{\text{BSS}}$ & $\text{GMGS}\uparrow$ & $\text{BSS}_{\ge \text{M}}\uparrow$ & $\text{TSS}_{\ge \text{M}}\uparrow$ & HM \\
        \hline \hline
        (i) &  & \checkmark & \checkmark & \checkmark & 0.399 & 0.414 & 0.367 & 0.407 \\
        (ii) & \checkmark & & \checkmark & \checkmark & 0.390 & 0.427 & 0.272 & 0.407 \\
        (iii) & \checkmark & \checkmark & & \checkmark & 0.286 & \textbf{0.551} & 0.308 & 0.377 \\
        (iv) & \checkmark & \checkmark & \checkmark & & 0.381 & 0.411 & 0.242 & 0.395 \\
        ours & \checkmark & \checkmark & \checkmark & \checkmark & \textbf{0.484} & 0.353 & \textbf{0.447} & \textbf{0.408} \\
        \hline
      \end{tabular}%
    
\end{table}

We conducted the following four ablation studies.

\begin{enumerate}[label=(\roman*)]
    \item To evaluate the contribution of the positional embedding, we trained the model without using it.
    \item To assess the impact of the IB loss ($L^{\text{IB}}_{\text{CE}}$), we removed it from the loss function.
    \item To examine the effect of using the Weighted BSS loss ($L^{'}_{\text{BSS}}$), we replaced it with the standard BSS loss that does not account for class-wise weighting.
    \item To investigate the role of the IB BSS loss ($L^{\text{IB}}_{\text{BSS}}$), we excluded it from the total loss.
\end{enumerate}

Table~\ref{tab:ablation} shows the results of the ablation studies.  
In model (i), the GMGS and $\text{BSS}_{\ge \text{M}}$ scores decreased by 0.085 and 0.080, respectively, indicating that the positional embedding contributes positively to the overall performance.

Furthermore, in models (ii), (iii), and (iv), the harmonic mean of GMGS and $\text{BSS}_{\ge \text{M}}$ declined compared with that in the full model, indicating that each of the loss components in the proposed method plays a meaningful role in improving the performance.


\begin{table}[tb]
  \centering
  \caption{Confusion matrix for the test set of the first fold.}
  \label{tab:confusion_matrix}
  \setlength{\tabcolsep}{8pt}
 \begin{tabular}{c|c||c c c c}
    \hline
    \multicolumn{2}{c||}{\multirow{2}{*}{}} & \multicolumn{4}{c}{Predicted flare class} \\
    \cline{3-6}
    \multicolumn{2}{c||}{} & O & C & M & X \\
    \hline\hline
    \multirow{4}{*}{Observed flare class} & O & 5336 & 471 & 92 & 69 \\
                                & C & 807 & 748 & 105 & 183 \\
                                & M & 139 & 130 & 85 & 64 \\
                                & X & 1 & 33 & 12 & 31 \\
    \hline
  \end{tabular}
\end{table}

Table~\ref{tab:confusion_matrix} shows the confusion matrix for the test set of the first fold.
From this result, we observe a tendency for the model to overpredict X-class flares, i.e., to classify non-X-class samples as X-class more frequently than other types of misclassification.
In this study, we define failure cases as samples whose predicted class does not match the ground-truth label.
Based on this definition, the numbers of failure cases in the first, second, and third folds of the time-series cross-validation were 2,106, 3,621, and 4,178, respectively.


\begin{table}[t]
  \centering
  \caption{Results of an error analysis of the test set using GMGS-Influence.}
  \label{tab:error}
  \setlength{\tabcolsep}{10pt}
  \begin{tabular}{ccc}
    \hline
    Observed class & Predicted class & GMGS-Influence \\
    \hline \hline
    C & O & 0.0741 \\
    O & C & 0.0433 \\
    M & O & 0.0195 \\
    C & X & 0.0163 \\
    O & M & 0.0129 \\
    \hline
  \end{tabular}
\end{table}

Table~\ref{tab:error} presents the results of an error analysis of the test set using GMGS-Influence~\cite{Kaneda_FlareTransformer_2022}.
We computed the impact of each failure case on the GMGS score using GMGS-Influence.
The GMGS-Influence corresponding to element $(i, j)$ in the confusion matrix is defined as
\begin{equation}
    \text{GMGS-Influence}_{ij} = \frac{c_{ij}(s_{ii}-s_{ij})}{N},
\end{equation}
where $c_{ij}$ denotes the $(i, j)$ element of the confusion matrix, $s_{ij}$ is the $(i, j)$ element of the GMGS scoring matrix, and $N$ is the total number of samples.
From Table~\ref{tab:error}, we observe that the mutual misclassifications between the C-class and O-class flares contribute most significantly to the degradation of the overall performance, suggesting that this confusion is a major bottleneck in the model accuracy.

\section{Conclusions}

In this study, we addressed the task of predicting the class of the largest solar flare expected to occur within the next 72 hours formulated as a multi-class classification problem.

The contributions of this study are summarized as follows.
\begin{itemize}
    \item We proposed the FLARE loss, which consists of 
    (i) the IB BSS loss, which incorporates influence-based weighting into the BSS loss;  
    (ii) the Weighted BSS loss, which applies class-wise weighting to the BSS loss; and  
    (iii) the IB loss, which suppresses the excessive influence of samples near the decision boundary.
    
    \item We performed 72-hour solar flare prediction using temporal embeddings derived from the 11-year solar activity cycle.
    
    \item The proposed method outperformed baseline methods in terms of GMGS and $\text{TSS}_{\ge \text{M}}$.
\end{itemize}

For future work, extending the sequence length of input images could help capture longer-term patterns in the solar activity cycle.
Additionally, using higher-resolution images may enable more advanced spatial modeling, potentially leading to further improvements in the prediction performance.


\subsection*{Acknowledgment}
\small
This work was partially supported by JSPS KAKENHI Grant Number 23K28168.

\bibliographystyle{splncs04}
\bibliography{reference}

\end{document}